\documentclass[runningheads]{llncs}
\usepackage{graphicx}
\graphicspath{{figures/}}
\usepackage{color}

\usepackage{caption}
\usepackage{subfig}
\usepackage{booktabs}
\usepackage{multirow}
\usepackage{comment}
\usepackage{hyperref}
\begin{document}
\title{A Benchmark dataset for predictive maintenance}
%
%
\author{
Bruno Veloso\inst{4,5}\orcidID{0000-0001-7980-0972} \and
Jo\~ao Gama \inst{1,3, 4}\orcidID{0000-0003-3357-1195} \and Rita P. Ribeiro \inst{2,3,4}\orcidID{0000-0002-6852-8077} \and 
Pedro M. Pereira\inst{4}}

\authorrunning{Veloso et al.}
%
\institute{
Faculdade de Economia
   \email{\{up201801860, jgama\}@fep.up.pt}
\and 
Faculdade de Ciências
   \email{rpribeiro@fc.up.pt}
\and 
University of Porto 
\and INESC TEC
    \email{\{bruno.miguel.veloso,pm.pereira.mail\}@gmail.com }
\and University Portucalense
}

\maketitle              

\begin{abstract}
The paper describes the {\tt MetroPT} data set, an outcome of a eXplainable Predictive Maintenance (XPM) project with an urban metro public transportation service in Porto, Portugal. The data was collected in 2022 that aimed to evaluate machine learning methods for online anomaly detection and failure prediction. By capturing several analogic sensor signals (pressure, temperature, current consumption), digital signals (control signals, discrete signals), and GPS information (latitude, longitude, and speed), we provide a dataset that can be easily used to evaluate online machine learning methods. This dataset contains some interesting characteristics and can be a good benchmark for predictive maintenance models.

\end{abstract}




\section*{Background \& Summary}
The occurrence of faults in public transport vehicles during their regular operation is a source of numerous damages, especially when they cause the interruption of the trip. The negative impacts affect not only the operator company but the clients, who are thereby disappointed with their expectations of transportation trust. In this context, the early detection of such faults can avoid the cancellation of trips and the withdrawal of service from the respective vehicle and thus is of enormous value. Only in 2017, more than 170 trips were canceled for this reason.

The Air Production Unit (APU) installed on the roof of Metro vehicles feeds units that perform different functions. One of these units is the secondary suspension, responsible for maintaining the height of the vehicle level regardless of the onboard number of passengers. The APU is a highly requested element on the vehicle throughout the day. The absence of redundancy causes its failure to result in the immediate removal of the train for repair. The failures typically are undetectable according to traditional maintenance criteria (predefined thresholds).

From the operational point of view, the objective of Predictive Maintenance is to reduce operational problems, reduce the number of unforeseen stops and the stopping time, and change the maintenance paradigm: from reactive to predictive.

The main goal of the {\tt MetroPT} dataset is to become a benchmark dataset for Predictive Maintenance.  
That is a real-world dataset, where the ground truth of anomalies is known from the company's maintenance reports.
It will allow fair comparisons between Machine Learning algorithms developed to detect anomalies based on sensor data collected as a continuous data flow.


\section*{Methods}

\begin{figure}[!ht]
    \centering
    \includegraphics[width=1\textwidth]{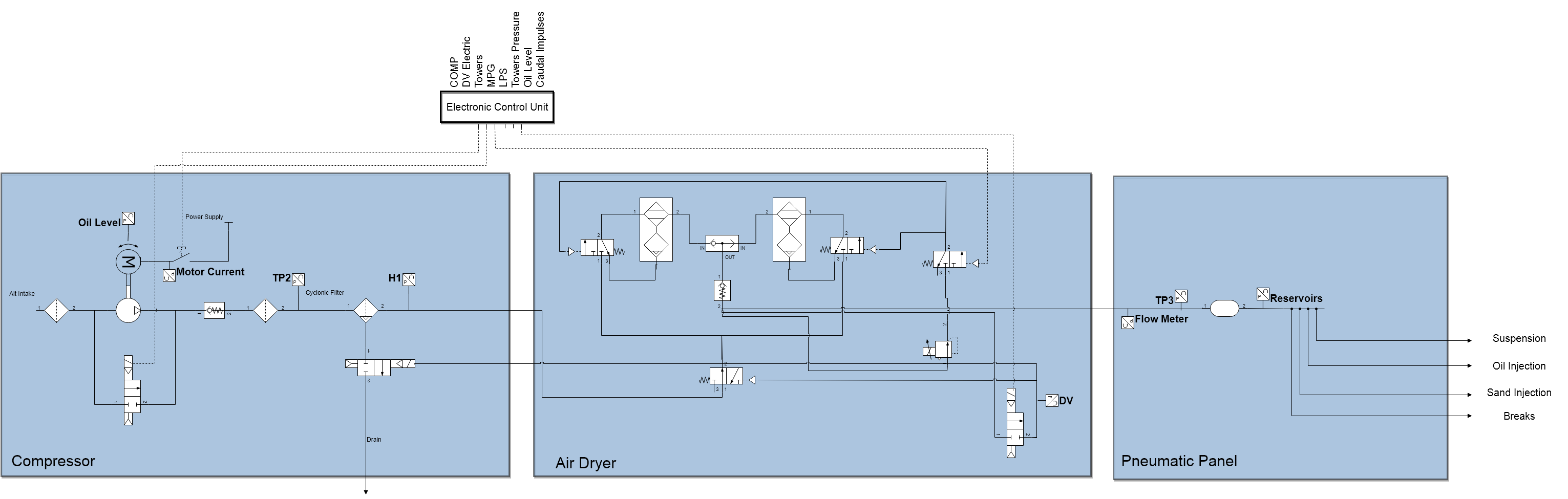}
    \caption{Air Producing Unit (APU)}
    \label{fig:apu}
\end{figure}

In the last few years, many works have been published about Predictive Maintenance (PdM) with the development of machine and deep learning techniques. 
A recent survey in Predictive Maintenance was published by \cite{Esteban2022}. It covers the main issues in data-driven PdM.
Another survey describing advances using machine learning and deep learning techniques for handling PdM in the railway industry was published by \cite{davari2021survey}. A recent manuscript by~\cite{gama2022data} identifies three key research lines for the PdM domain: failure prediction, remaining useful life (RUL), and root cause analyses (RCA) as some exciting topics that will attract the focus of researchers. 
In fact the final goal of PdM consists of elaborating a maintenance plan when a failure is identified. Identifying the components involved in the failure and the severity of the failure are relevant informations for the recovering plan.

Regarding the present dataset, two recent works try to solve the failure prediction. In the first, presented by \cite{barros2020failure}, the authors constructed a rule-based system to produce some alerts about the state of the compressor. The second work, presented by \cite{davari2021predictive}, explores the usage of deep learning auto encoders to produce alerts. In both cases, the results are satisfactory, but there is a vast space to improve accuracy and explanation.






 

\section*{Data Records}
\begin{figure}[!h]
    \centering
    \includegraphics[width=1.0\textwidth]{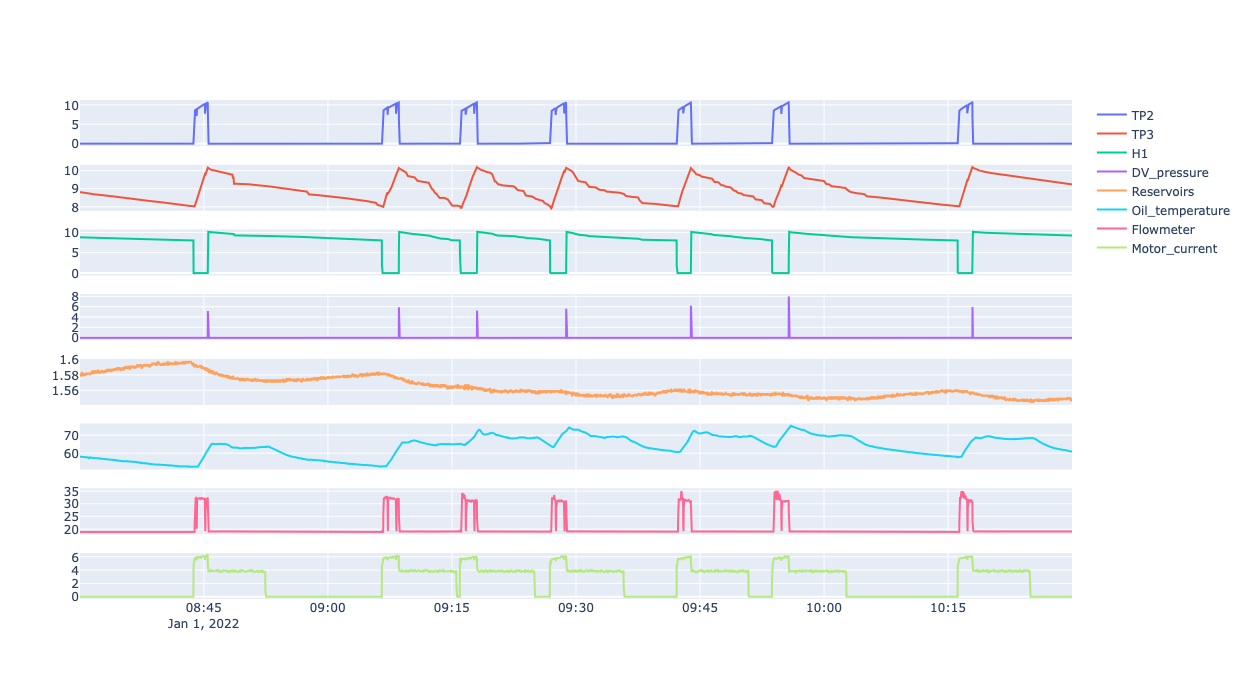}
    \caption{Analog Sensors}
    \label{fig:analog}
\end{figure}

\begin{figure}[!h]
    \centering
    \includegraphics[width=1.0\textwidth]{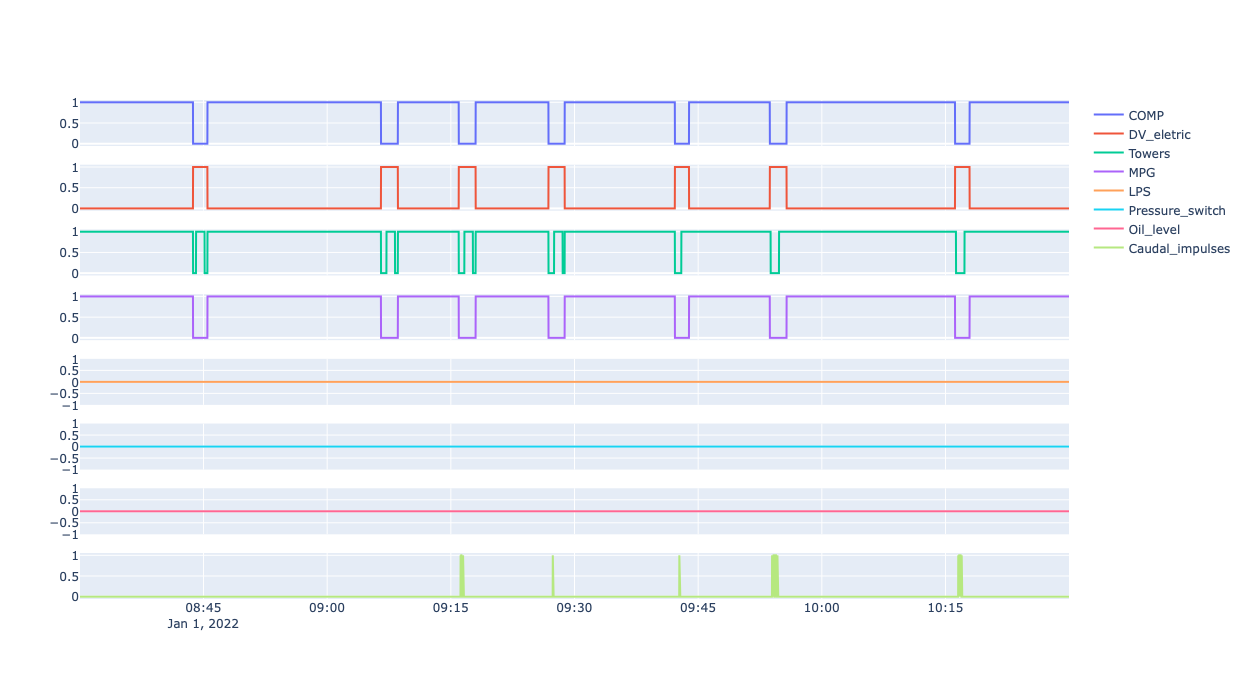}
    \caption{Digital Sensors}
    \label{fig:digital}
\end{figure}

\begin{figure}[!h]
    \centering
    \includegraphics[width=1.0\textwidth]{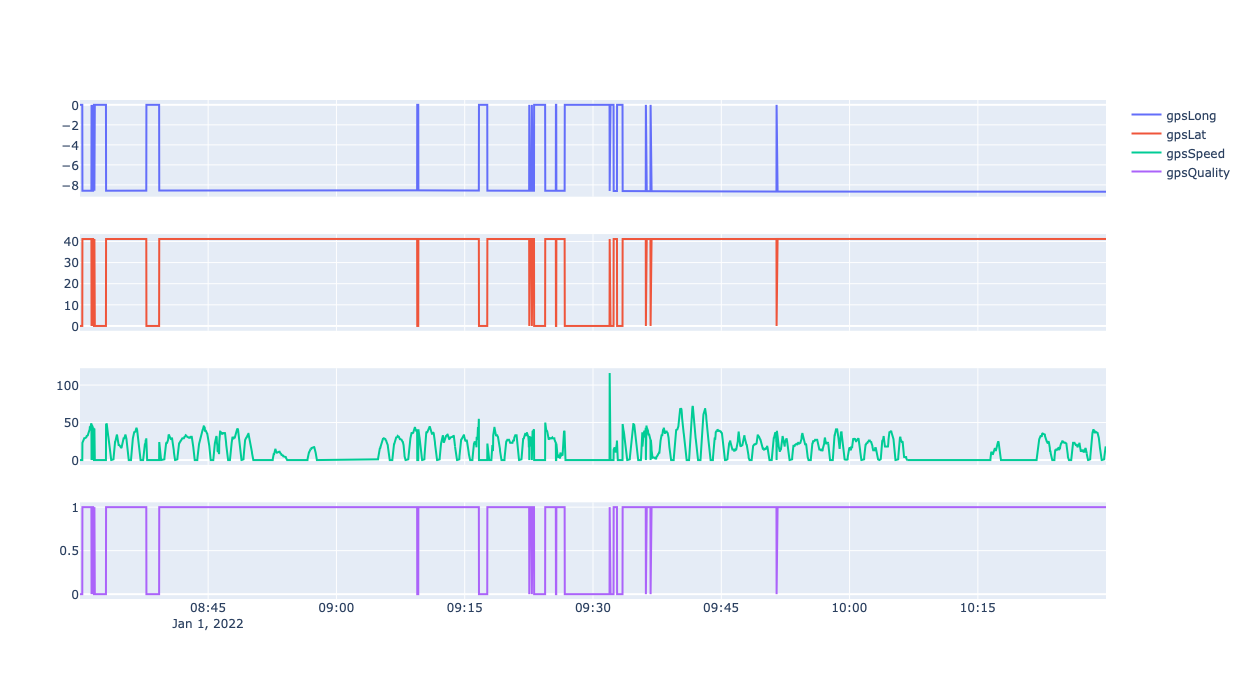}
    \caption{GPS Information}
    \label{fig:gps}
\end{figure}

The signal acquisition system installed in one APUs of vehicle (APU01) collects data from eight analogue sensors (see Figure \ref{fig:analog}) (pressure, temperature and electric current consumed, placed in different components of the APU - Figure \ref{fig:apu}) as well as eight digital signals collected directly from the APU (see Figure \ref{fig:digital}) and GPS information (see Figure \ref{fig:gps}. 

The data acquisition rate is 1Hz, and the information is sent to the remote server every 5 minutes using the GSM network. The data collection of the two units began on 12 March 2020 and is continuously operational to date. Every day, for each APU, a report is generated with the information of the sensor signals.

The considered analogue sensors (Figure \ref{fig:analog}) are the following:
\begin{itemize}
    \item {\tt TP2} \cite{pt5414} - Measures the pressure on the compressor.
    \item {\tt TP3} \cite{pt5414} - Measures the pressure generated at the pneumatic panel. 
    \item {\tt H1} \cite{pt5414} - This valve is activated when the pressure read by the pressure switch of the command is above the operating pressure of 10.2 bar.
    \item {\tt DV pressure} \cite{pt5414} - Measures the pressure exerted due to pressure drop generated when air dryers towers discharge the water. When it is equal to zero, the compressor is working under load. 
    \item {\tt Motor Current} \cite{at-b420l} - Measure the motor's current, which should present the following values: (i) close to 0A when the compressor turns off; (ii) close to 4A when the compressor is working offloaded; and (iii) close to 7A when the compressor is operating under load. 
    \item {\tt Oil Temperature} \cite{tc12-m} - Measure the temperature of the oil present on the compressor.
\end{itemize}

The digital sensors (Figure \ref{fig:digital}) only assume two different values: zero when inactive or one when a specific event activates them. The considered digital sensors were the following:
\begin{itemize}
    \item COMP - The electrical signal of the air intake valve on the compressor. It is active when there is no admission of air on the compressor, meaning that the compressor turns off or working offloaded. 
    \item DV electric - electrical signal that commands the compressor outlet valve. When it is active, it means that the compressor is working under load, and when it is not active, it means that the compressor is off or working offloaded.
    \item TOWERS - Defines which tower is drying the air and which tower is draining the humidity removed from the air. When it is not active, it means that tower one is working, and when it is active, it means that tower two is working.
    \item MPG - Is responsible for activating the intake valve to start the compressor under load when the pressure in the APU is below 8.2 bar. Consequently, it will activate the sensor COMP, which assumes the same behaviour as MPG sensor. 
    \item LPS - Is activated when the pressure is lower than 7 bars.
    \item Oil Level - Detects the oil level on the compressor and is active (equal to one) when the oil is below the expected values.
\end{itemize}

Regarding the GPS Information, the train was equipped with a secondary GPS antenna to collect: latitude, longitude, speed and signal quality.

\begin{figure}[!h]
    \centering
    \includegraphics[width=1.0\textwidth]{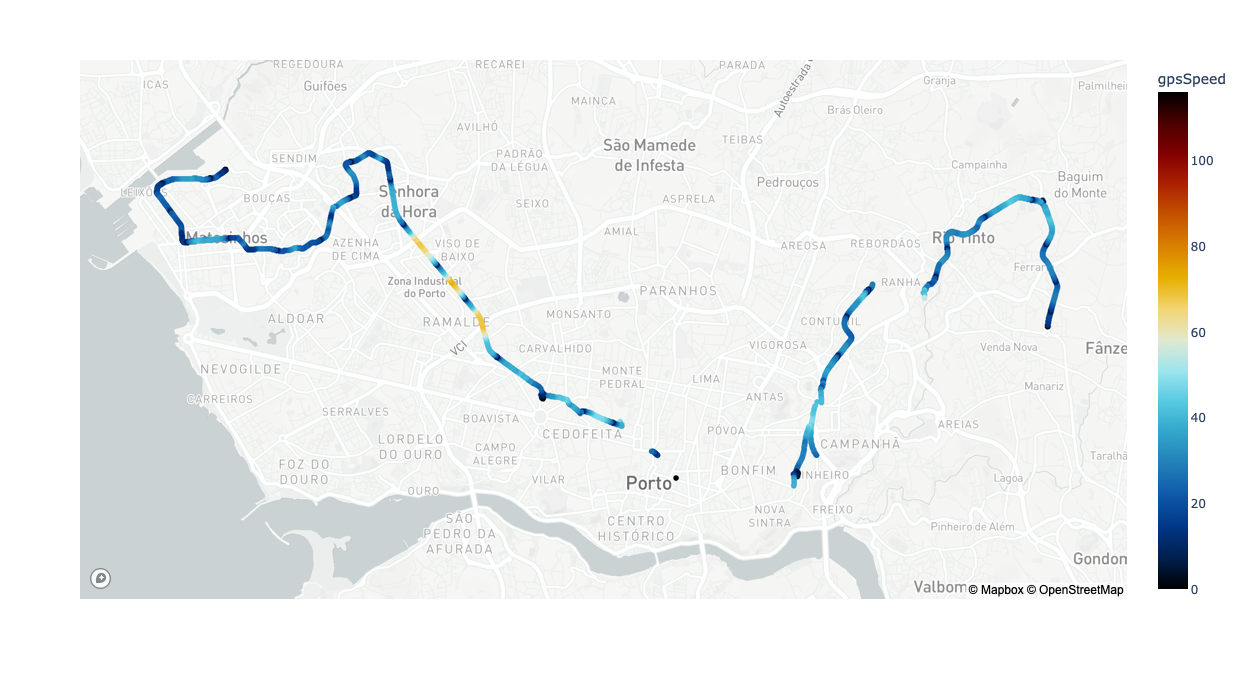}
    \caption{GPS Information. The train trajectory is interrupted when the train is inside a tunnel, due to loss of the GPS signal.}
    \label{fig:map}
\end{figure}

The data set available was collected from January to June 2022. It contains a single file with all variables and GPS coordinates, with almost $10 979 547$ data points collected from the air compressor.



\section*{Technical Validation}

The ground truth was provided by the company using maintenance reports. According to the reported information, the dataset has three catastrophic failures. Two failures are related to air leaks in the system, and another is an oil leak.
\begin{itemize}
\item Regarding the air leaks, the first one \ref{figs2a} is provoked by a malfunction on the pneumatic pilot valve that opens the drain pipes during the operation of the compressor.

 \begin{figure}[!h] 
         \centering
         \includegraphics[width=\textwidth]{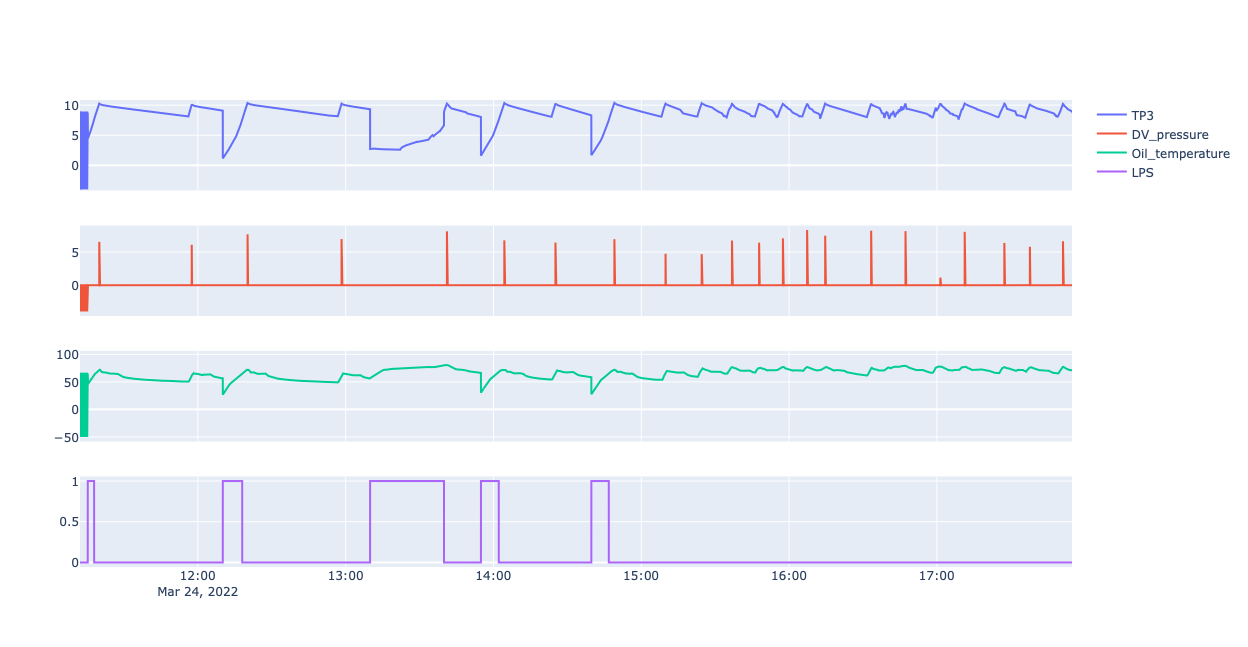}
         \caption{APU Control Failure - Air escaping by Drain Pipes}
         \label{figs2a}
     \end{figure}

\item The second problem \ref{figs2b} was an air leak on a pipe that feeds several clients on the systems, such as breaks, suspension, etc. In the first case, the train recovered from the malfunction. In the second case, the train needed to move to the maintenance building.

\begin{figure}[!h]
         \centering
         \includegraphics[width=\textwidth]{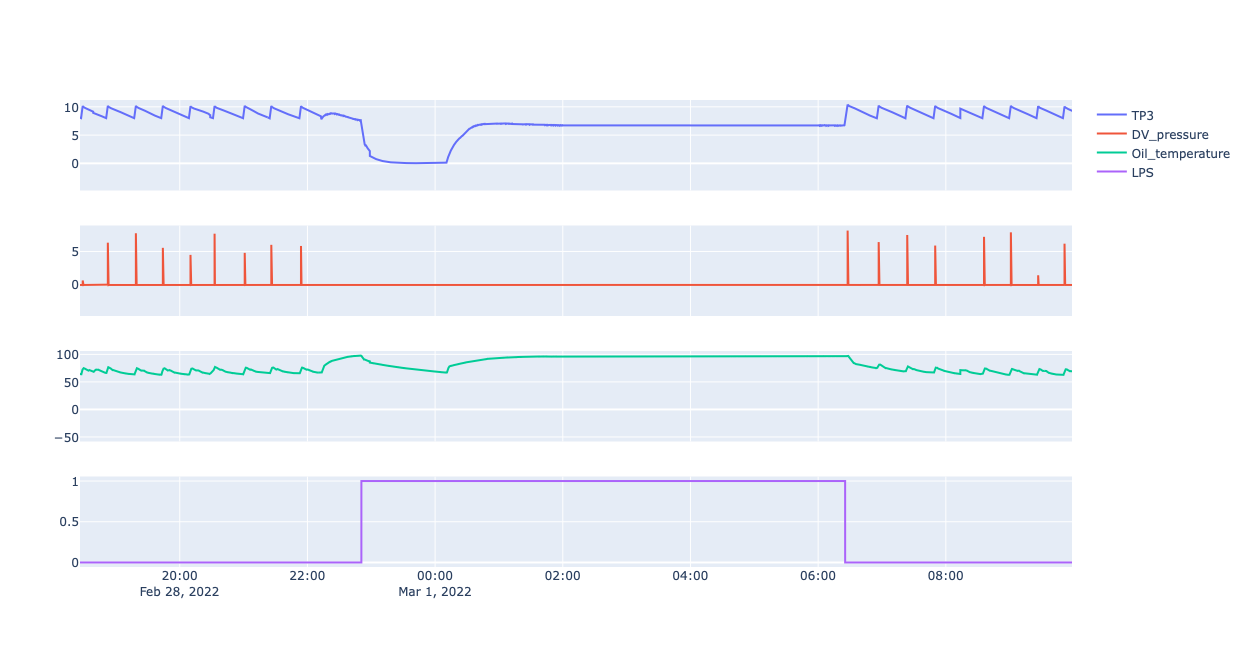}
         \caption{Air Leak on clients}
         \label{figs2b}
     \end{figure}
     
\item Regarding the oil leak \ref{figs2c}, due to hardware design, there was not any signal system related to oil to warn the train driver, the oil leak provoked severe damage to the engine of the compressor, and subsequentially, due to the inoperable compressor, it was observed a drop on the air pressure and the train needed to be removed from the tracks.

        \begin{figure}[!h]
         \centering
         \includegraphics[width=\textwidth]{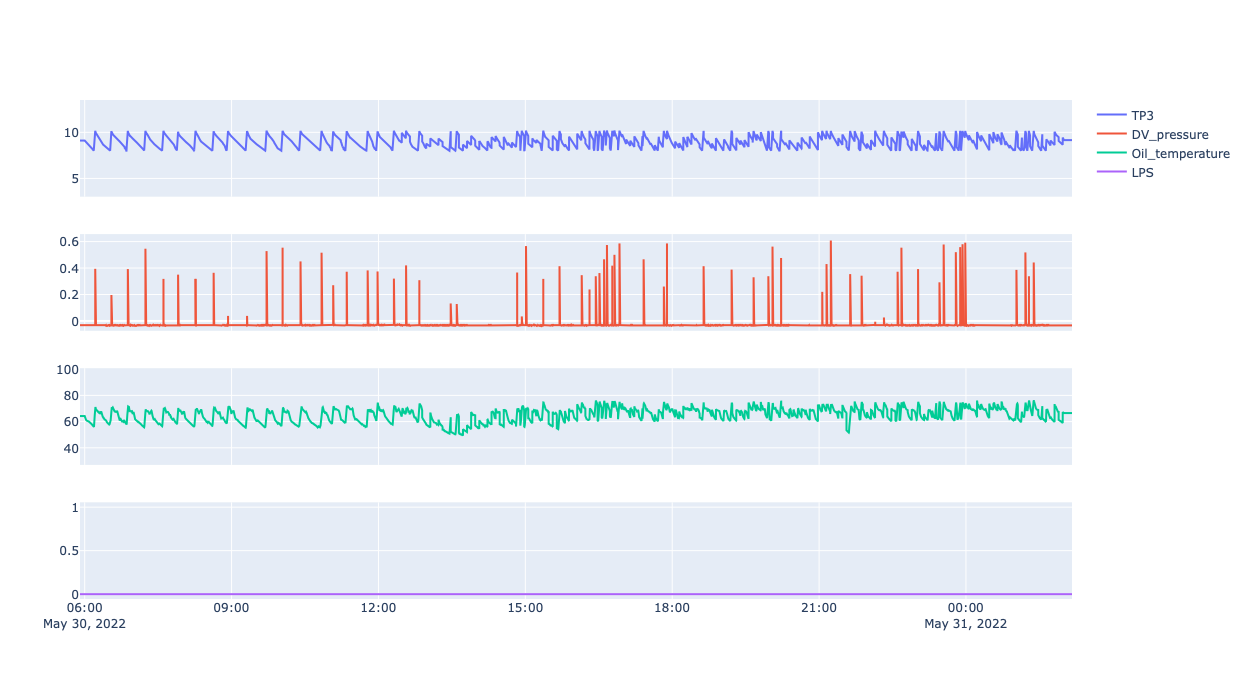}
         \caption{Oil Leak on Compressor}
         \label{figs2c}
     \end{figure}
\end{itemize}

\begin{table}[!h]
\begin{tabular}{llccl}
    \toprule
     Nr. & Type & Component & Start &End \\
     \midrule
         1 & Air Leak & Air Dryer &28-2-22 21:53 & 1-03-22 02:00\\
         2 & Air Leak & Clients & 23-3-22 14:54 & 23-03-22 15:24\\
         3 & Oil Leak & Compressor & 30-5-22 12:00 & 02-06-22 06:18\\
     \bottomrule
\end{tabular}
\caption{Failures reported on Maintenance Reports.}
\end{table}

\begin{figure}[!h]
    \centering
    \includegraphics[width=1.0\textwidth]{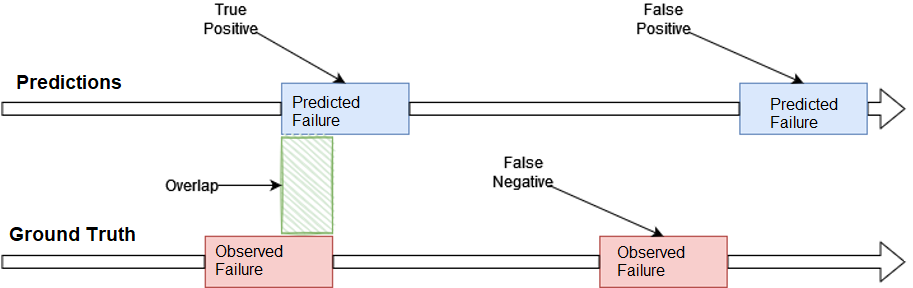}
    \caption{Evaluation Protocol}
    \label{fig:eval}
\end{figure}

The dataset can be used for two primary purposes: i) Predicting failures and ii) Identifying the components involved in the failure.
For the first task predicting failures, the goal is to predict when it starts and the duration of the failure. 
For validation purposes, a failure is a time interval: start-end. We use the the following evaluation protocol:
\begin{itemize}
    \item \textbf{Goal}: is to minimize the number of false positives and false negatives (\# FP + \# FN) (see Figure \ref{fig:eval})
    \item \textbf{Requirement}: from the company is to detect the failure at least two hours before the train becomes non-operational to remove it from the tracks safely.
\end{itemize}

Our goal is to discover the problems as early as possible after it manifests, i.e., to increase the overlap between the prediction and the ground truth. The second task is to identify the type of failure and in which component the failure occurs. Finally, it is crucial to compute the remaining useful life of the components to help the management team when they need to remove the train without provoking disruptions to the service.

\section*{Acknowledgements}

This work was supported by the CHIST-ERA grant CHIST-ERA-19-XAI-012, and project CHIST-ERA/0004/2019 funded by FCT.





\bibliography{sample}
\bibliographystyle{splncs04}

\end{document}